\documentclass[lettersize,journal]{IEEEtran}
\usepackage{amsmath,amsfonts}
\usepackage{algorithmic}
\usepackage{algorithm}
\usepackage{array}
\usepackage[caption=false,font=normalsize,labelfont=sf,textfont=sf]{subfig}
\usepackage{textcomp}
\usepackage{url}
\usepackage{verbatim}
\usepackage{graphicx}
\usepackage{cite}
\usepackage{booktabs}
\usepackage{graphicx}
\usepackage{caption}
\usepackage{subcaption}
\usepackage{hyperref}
\begin{document}
	
	\title{Inner-IoU: More Effective Intersection over Union Loss with Auxiliary Bounding Box}
	
	\author{Hao Zhang, Cong Xu, Shuaijie Zhang}
	\maketitle
	
	\begin{abstract}
		\textbf{\bfseries{With the rapid development of detectors, Bounding Box Regression (BBR) loss function has constantly updated and optimized. However, the existing IoU-based BBR still focus on accelerating convergence by adding new loss terms, ignoring the limitations of IoU loss term itself. Although theoretically IoU loss
		can effectively describe the state of bounding box regression,
		in practical applications, it cannot adjust itself according to
		different detectors and detection tasks, and does not have strong
		generalization. Based on the above, we first analyzed the BBR model and concluded that distinguishing different regression samples and using different scales of auxiliary bounding boxes to calculate losses can effectively accelerate the bounding box regression process. For high IoU samples, using smaller auxiliary bounding boxes to calculate losses can accelerate convergence, while larger auxiliary bounding boxes are suitable for low IoU samples. Then, we propose Inner-IoU loss, which calculates IoU loss through auxiliary bounding boxes. For different datasets and detectors, we introduce a scaling factor ratio to control the scale size of the auxiliary bounding boxes for calculating losses. Finally, integrate Inner-IoU into the existing IoU-based loss functions for simulation and comparative experiments. The experiment result demonstrate a further enhancement in detection performance with the utilization of the method proposed in this paper, verifying the effectiveness and generalization ability of Inner-IoU loss. Code is available at \url{https://github.com/malagoutou/Inner-IoU}.}}
	\end{abstract}
	
	\begin{IEEEkeywords}
		\textbf{simple principle, easy to use and strong generalization.}
	\end{IEEEkeywords}
	
	\section{Introduction}
	\IEEEPARstart{O}{bject} detection is a fundamental task in computer vision, which includes object classification and localization. The bounding box regression loss function is an important component of the detector localization branch and the positioning accuracy of the detector largely depends on bounding box regression, which plays an irreplaceable role in  the current detectors. 
	\begin{figure}[!t]
		\centering
		\includegraphics[width=\linewidth]{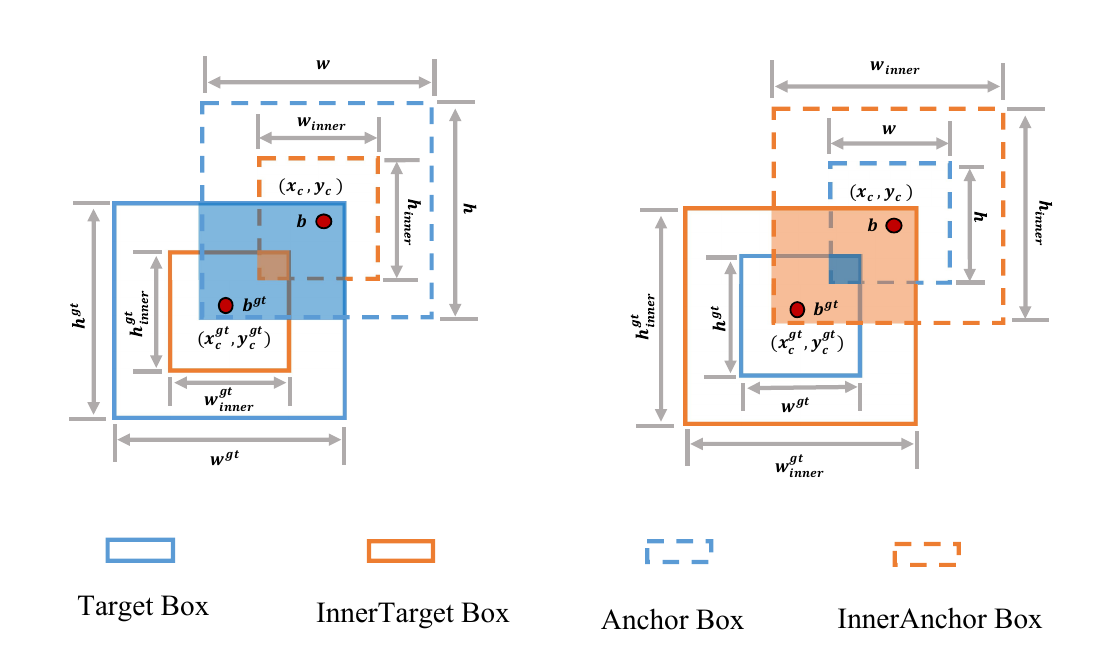}
		\caption{Description of Inner-IoU }
		\label{fig_1} 
	\end{figure}
	In the BBR, IoU loss\cite{ref1} can accurately describe the degree of matching between the predicted bounding box and the GT box, ensuring that the model can learn the position information of the target during the training process.
	As an essential part  of the existing mainstream bounding box regression loss functions, IoU is defined as follows:
	\begin{equation} 
		IoU=\displaystyle\frac{\left\vert B\cap B^{gt} \right\vert}{\left\vert B\cup B^{gt} \right\vert}
	\end{equation}
	\par $B$ and $B^{gt}$  represent the predicted box and the GT box, respectively. After defining IoU, its corresponding loss can be defined as follows:
	\begin{equation} 
		L_{IoU}=1-IoU 
	\end{equation}
	\par So far, IoU-based loss functions have gradually become mainstream and dominate. Most of the existing methods are based on IoU and further adding new loss terms. For example, GIoU\cite{ref2} was proposed to solve the gradient vanishing problem when the overlap area between anchor box and GT box is 0.The GIoU loss function\cite{ref2} is defined as follows, where $C$ is the smallest box covering $B$ and $B^{gt}$:  	
	\begin{equation} 
		L_{GIoU}=1-IoU+\displaystyle\frac{\left\vert C-B\cap B^{gt} \right\vert}{\left\vert C \right\vert}
	\end{equation} 
	Compared with GIoU, DIoU loss\cite{ref3} function proposed to add a new distance loss term on the basis of IoU, mainly by minimizing the normalized distance between the center points of the two bounding boxes. This allows it to achieve faster convergence and better performance. It is represented as follows:
	\begin{equation}
		L_{DIoU}=1-IoU+\displaystyle\frac{\rho^{2}(b,b^{gt})}{c^{2}}
	\end{equation}
	Where $b$ and $b^{gt}$  are the center points of ${B}$ and $B^{gt}$ respectively, $\rho\left(\cdot\right) $  refers to the Euclidean distance, where $c$ is the diagonal of the minimum bounding box.
	\begin{figure*} [h]
		\centering
			\includegraphics[scale=0.3]{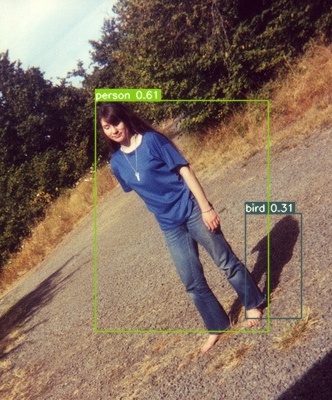}
		\hspace{0.1cm}
		\vspace{0.4cm}
			\includegraphics[scale=0.3]{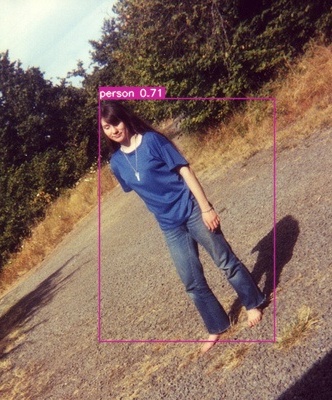}
		\hspace{0.1cm}
			\includegraphics[scale=0.3]{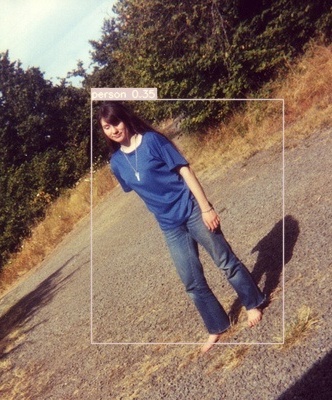}
		\hspace{0.1cm}
			\includegraphics[scale=0.3]{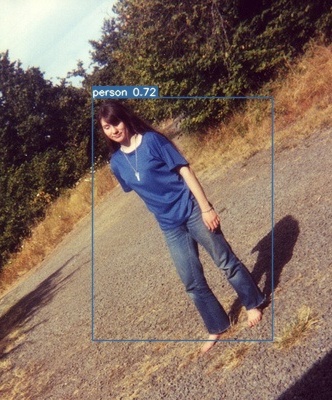}
		\\
		\centering
			\includegraphics[scale=0.3]{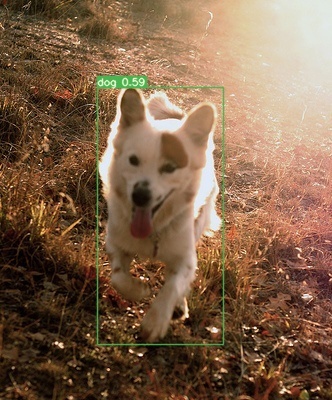} 
		\hspace{0.05cm}
			\includegraphics[scale=0.3]{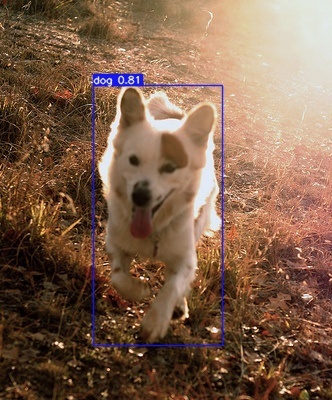}
		\hspace{0.1cm}
			\includegraphics[scale=0.3]{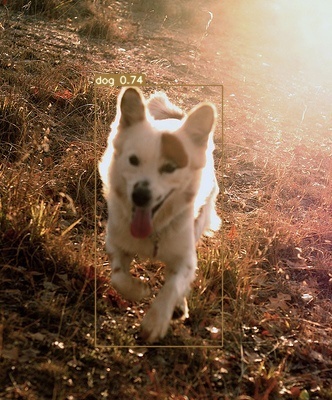}
		\hspace{0.1cm}
			\includegraphics[scale=0.3]{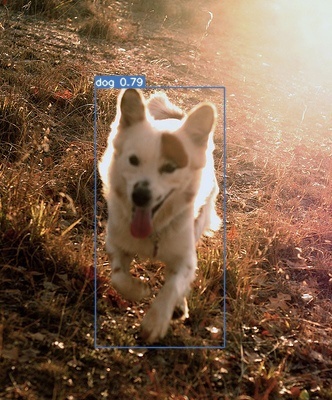}
		\caption{Detection examples on the test set of PASCAL VOC 2007 using YOLOv7-tiny by $L_{CIoU}$ and $L_{Inner-CIoU}$. From left to right, they represent the CIoU method, Inner-CIoU (ratio=0.7), Inner-CIoU (ratio=0.75) and Inner-CIoU (ratio=0.8). }
		\label{fig_2} 
	\end{figure*}
	\par The CIoU loss\cite{ref3} further considered the shape loss and added a shape loss term on the basis of DIoU loss. It is represented as follows:
	\begin{equation}
		L_{CIoU}=1-IoU+\displaystyle\frac{\rho^{2}(b,b^{gt})}{c^{2}}+\alpha v
	\end{equation}
	where $\alpha$ is a positive trade-off parameter:
	\begin{equation}
		\alpha=\displaystyle\frac{v}{(1-IoU)+v}
	\end{equation}
	where $v$ measures the consistency of aspect ratio:
	\begin{equation}
		v=\displaystyle\frac{4}{\pi^{2}}(arctan\displaystyle\frac{w^{gt}}{h^{gt}}-arctan\displaystyle\frac{w}{h})^{2}
	\end{equation}
	$w^{gt}$ and $h^{gt}$ denote the width and height of target box,$w$ and $h$ denote the width and height of predicted box.When the aspect ratio of target box and predicted box is the same, CIoU will degrade to DIoU.
	\par Compared with DIoU, the EIoU loss\cite{ref4} directly minimizes the normalized difference of the target box's and anchor box's width $(w,w^{gt})$, height $(h,h^{gt})$ and central location $(b,b^{gt})$.The EIoU loss function\cite{ref4} is defined as follows:
	\begin{equation}
		L_{EIoU}=1-IoU+\displaystyle\frac{\rho^{2}(b,b^{gt})}{c^{2}}+\displaystyle\frac{\rho^{2}(w,w^{gt})}{(w^{c})^{2}}+\displaystyle\frac{\rho^{2}(h,h^{gt})}{(h^{c})^{2}}
	\end{equation}
	 $w^{c}$ and $h^{c}$are the width and height of the minimum bounding box covering target box and predicted box. 
	\par The recent SIoU loss\cite{ref5} has taken into account the influence of the angles between anchor box and GT box on bounding box regression based on previous research and introduced angle loss into the bounding box regression loss function. It is defined as follows:
	\begin{equation}
		L_{SIoU}=1-IoU+\displaystyle\frac{(\Delta+\Omega)}{2}
	\end{equation}
	\par The angle loss represents the minimum angle between the central point's connection of the GT box and the anchor box:
	\begin{equation}
		\Lambda=sin(2sin^{-1}\displaystyle\frac{min(\left\vert x_{c}^{gt}-x_{c} \right\vert,\left\vert y_{c}^{gt}-y_{c} \right\vert)}{\sqrt{ (x_{c}^{gt}-x_{c})^{2}+(y_{c}^{gt}-y_{c})^{2}}+\in})
	\end{equation}
	\par This term aims to bring the anchor box to the nearest coordinate axis and consider whether to approach the X-axis or Y-axis preferentially according to the change of angle.When the angle value is $45^{\circ}$, $\Lambda$ = 1. When the central points are aligned on the X-axis or Y-axis, $\Lambda$ = 0. 
	\par The distance loss is redefined after taking the angle cost into account as follows: 
	\begin{equation}
		\Delta=\frac{1}{2}\sum_{t=w,h}(1-e^{-\gamma\rho_{t}}),\gamma=2-\Lambda
	\end{equation}
	\begin{equation}
		\left\{
		\begin{aligned}
			\displaystyle\rho_{x} & = & (\frac{b_{x}-b_{x}^{gt}}{w^{c}})^{2} \\
			\rho_{y} & = & (\frac{b_{y}-b_{y}^{gt}}{h^{c}})^{2} \\
		\end{aligned}
		\right.
	\end{equation}
	\par The shape loss mainly describes the size difference between the GT box and the anchor box, it is defined as follows:
	\begin{equation}
		\Omega=\frac{1}{2}\sum_{t=w,h}(1-e^{\omega_{t}})^{\theta},\theta=4
	\end{equation}
	\begin{equation}
		\left\{
		\begin{aligned}
			\displaystyle\omega_{w} & = & \frac{\left\vert w - w_{gt} \right\vert}{max(w,w_{gt})} \\
			\displaystyle\omega_{h} & = & \frac{\left\vert h - h_{gt} \right\vert}{max(h,h_{gt})} \\
		\end{aligned}
		\right.
	\end{equation}
	\par The value of $\theta$ determines the importance of the cost of shape.The range of this parameter is from 2 to 6.
	\par Although the above bounding box regression loss functions can effectively accelerate convergence and improve detection performance by adding new geometric constraints to the IoU loss functions. They did not consider the rationality of IoU loss itself, which largely determines the quality of the detection results. To compensate for this deficiency, we propose Inner-IoU loss, which is calculated using auxiliary bounding boxes to accelerate regression without adding any new loss terms.
	\par The main contributions of this article are as follows:
	\par$\bullet$ We analyze the process and patterns of bounding box regression and based on the inherent characteristics of the bounding box regression problem, propose using smaller auxiliary bounding box calculation losses during model training to have a gain effect on the regression of high IoU samples, while low IoU samples have the opposite effect.
	\par$\bullet$ We propose Inner-IoU Loss, which uses scale factor ratio control to generate auxiliary bounding box of different scales for calculating losses. Applying it to existing IoU based loss functions can achieve faster and more effective regression results.
	\par$\bullet$ We conducted a series of simulation and comparative experiments and the experimental results showed that the detection performance and generalization of our method are superior to existing methods and for datasets of different pixel sizes it achieved sota.
		\begin{figure}[!htbp]
		\centering
		\subfloat[]{
		\centering
		\includegraphics[width=1.65in]{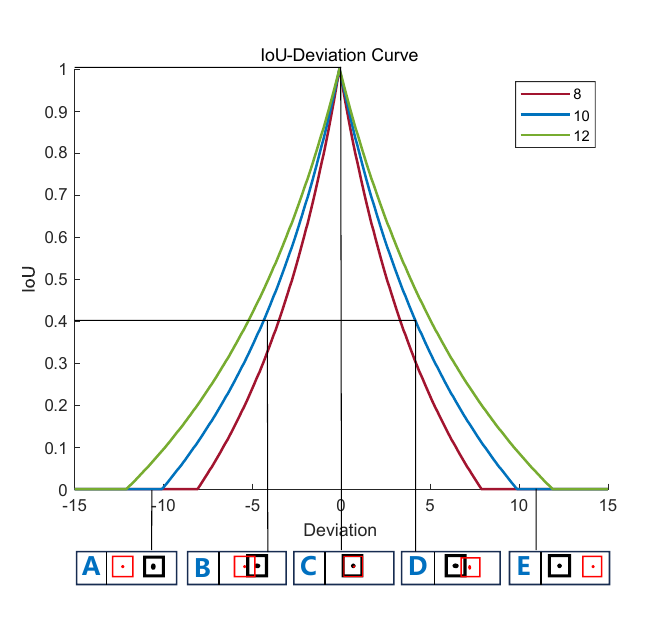}}
		\hfill 
		\subfloat[]{	
		\centering
		\includegraphics[width=1.65in]{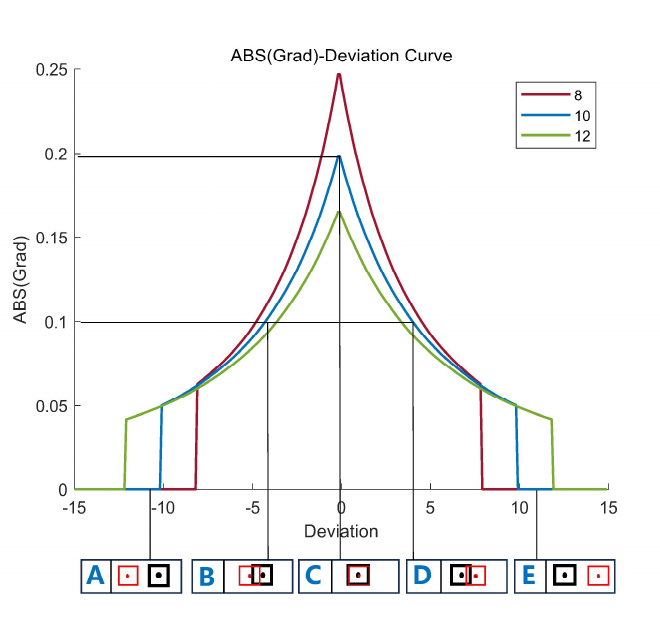}} 
		\caption{Regression change curves for different scale bounding boxes(a) IoU-Deviation Curve(b) ABS(Grad)-Deviation Curve}
		\label{fig_3}
	\end{figure}
	\begin{figure*} [t!]
		\centering
		\includegraphics[width=2in]{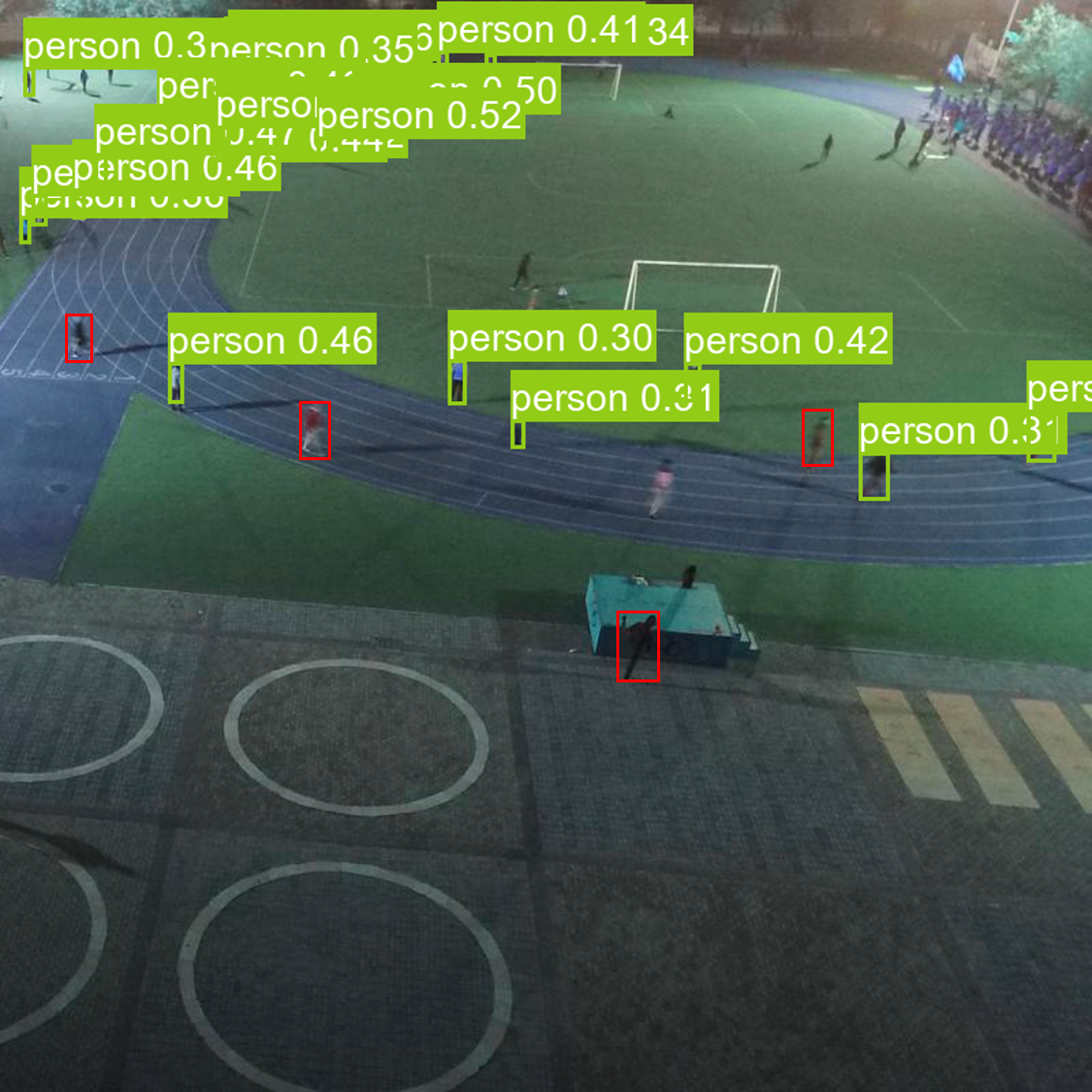}
		\hspace{.15in}
		\vspace{.15in}
		\includegraphics[width=2in]{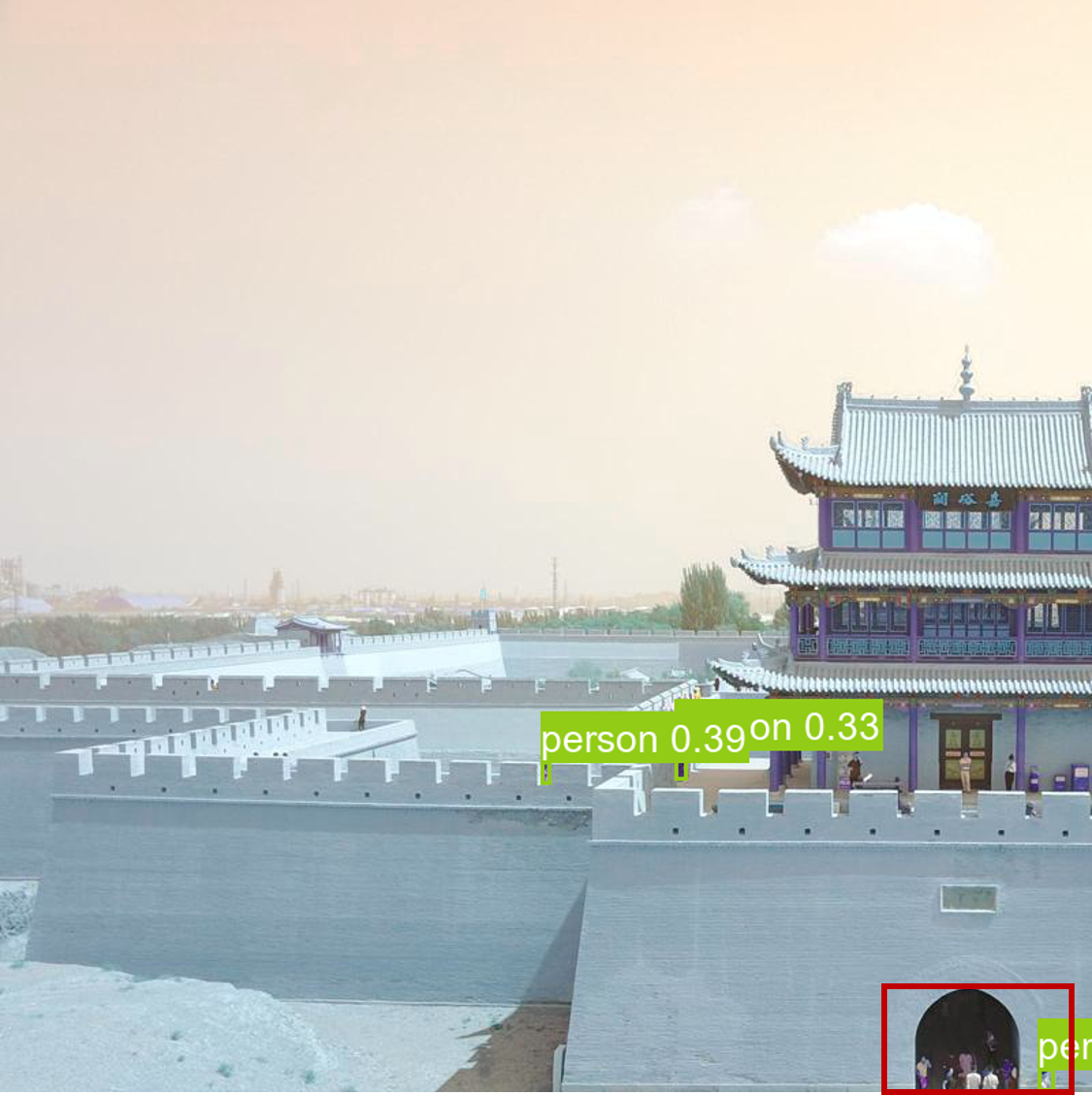}
		\hspace{.15in}
		\includegraphics[width=2in]{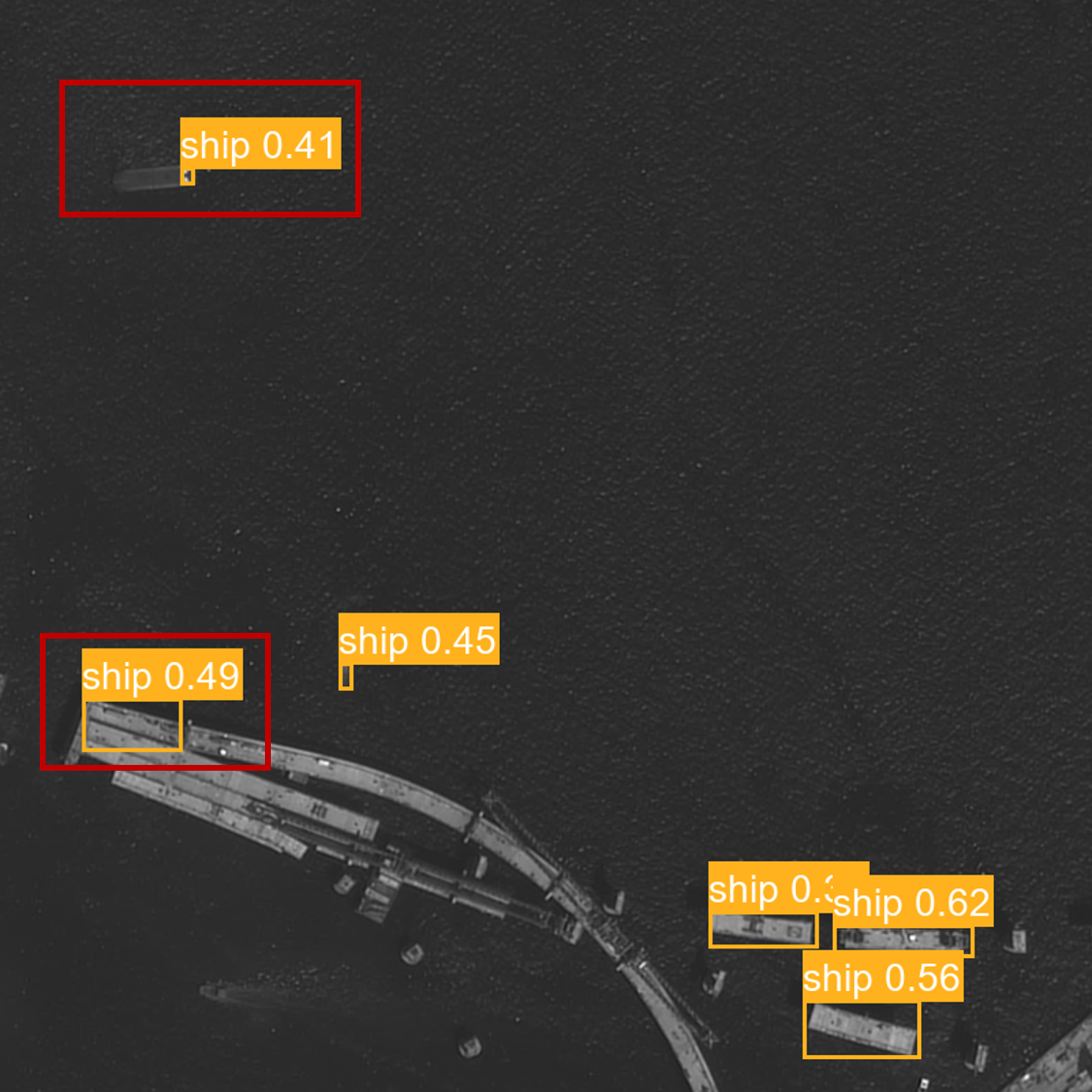} 
		\\
		\includegraphics[width=2in]{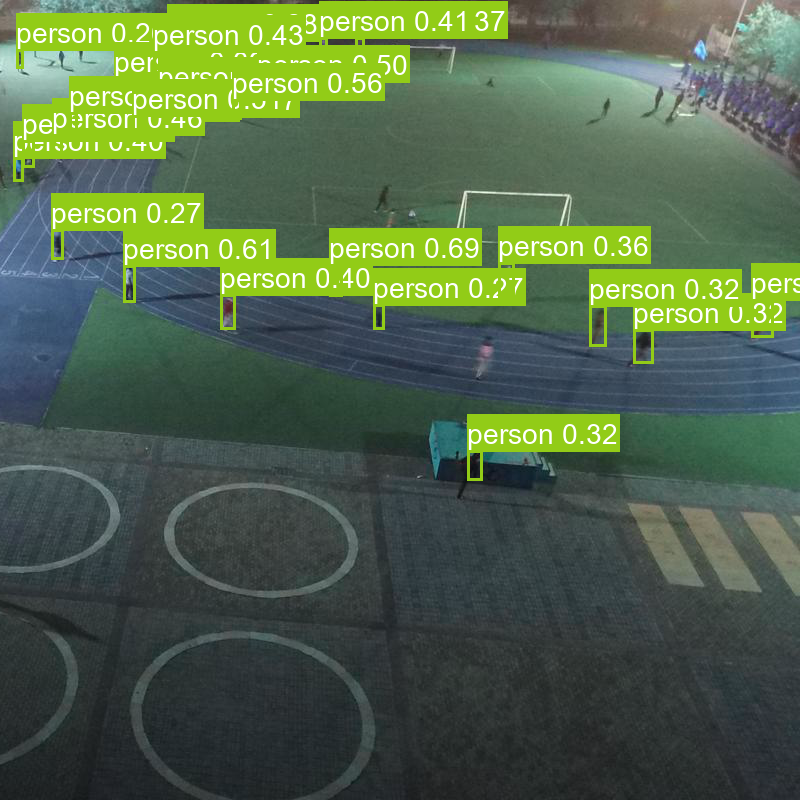}
		\hspace{.15in}
		\includegraphics[width=2in]{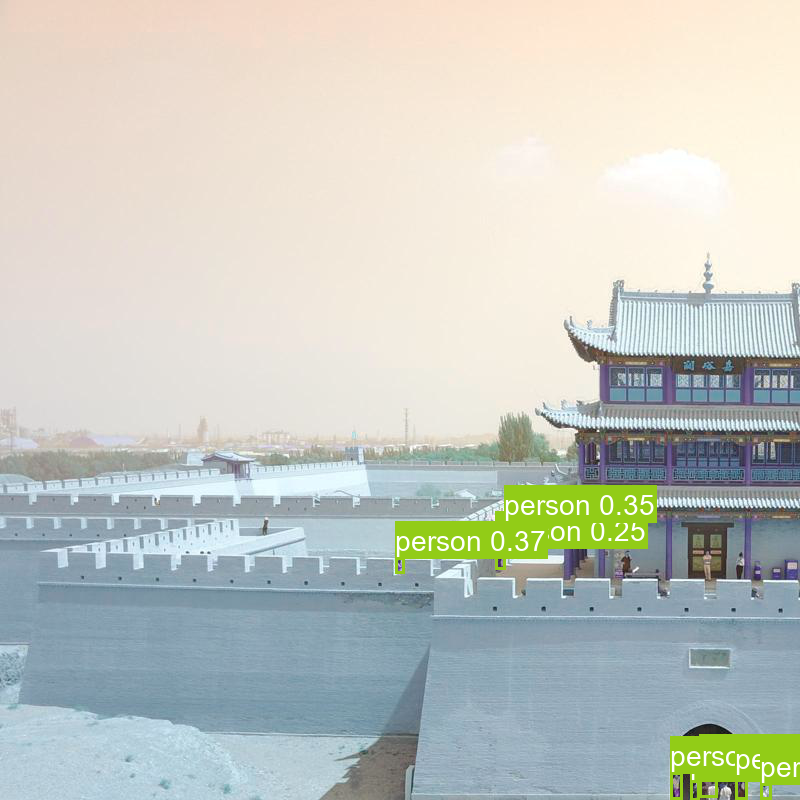} 
		\hspace{.13in}
		\includegraphics[width=2in]{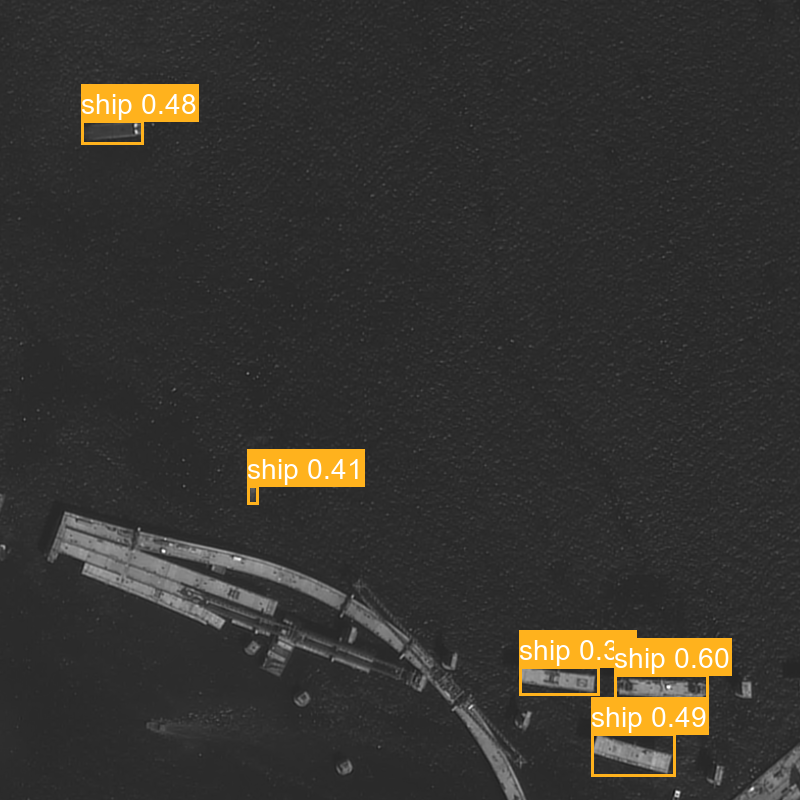}
		\caption{Detection examples on the test set of AI-TOD using YOLOv5s by $L_{SIoU}$ (first row) and $L_{Inner-SIoU}$(second row). }
		\label{fig_8} 
	\end{figure*}
	\section{Related work}
	\subsection{Object Detection}
	For object detection, it can be divided into anchor based and anchor free detection algorithms based on whether an anchor is generated. Anchor based algorithms include Faster R-CNN\cite{ref14}, YOLO (You Only Look Once) series\cite{ref10,ref11,ref12,ref13}, SSD (Single Shot MultiBox Detector)\cite{ref15} and RetinaNet\cite{ref16} Anchor free detection algorithms include CornerNet\cite{ref17}, CenterNet\cite{ref18} and FCOS (Fully Convolutional One Stage Object Detection)\cite{ref19}. In these object detection algorithms, the bounding box regression loss function plays a crucial role, enabling the detector to accurately locate the target and improves the detection accuracy of the detection algorithm.
	
	\subsection{Bounding Box Regression Losses}
	At first,  $\textit{l}_n$-norm loss\cite{ref14} is proposed as a bounding box regression loss, which is very sensitive to changes in bounding box scale. Subsequently, in order to compensate for this deficiency, IoU loss\cite{ref1} was proposed to replace $\textit{l}_n$-norm loss, compared to the $\textit{l}_n$-norm loss, the regression results of the IoU loss\cite{ref1} prediction box are more accurate. However, IoU loss cannot solve the gradient vanishing problem of non overlapping samples and GIoU loss\cite{ref2} compensates for this defect by introducing a minimum bounding box. The DIoU loss\cite{ref3} adds a distance constraint and the normalized distance between the central points of the prediction box and the GT box is added to the IoU loss as a new loss term, which improves convergence speed and position accuracy. The CIoU loss\cite{ref3} further considers the impact of shape similarity on bounding box regression, and adds a shape loss term to the DIoU loss. The EIoU loss\cite{ref4} uses focal loss to solve the problem of sample imbalance during the training process and redefines shape loss, further improving the detection effect. The latest SIoU\cite{ref5} loss adds the angle between the prediction box and the GT box as a new constraint to the bounding box regression loss, achieving the fastest convergence result. Compared with the above algorithms, we propose that Inner-IoU loss can further improve the convergence speed.
	\section{Method}
	\subsection{Bounding Box Regression Mode Analysis}
	\par The IoU loss function has a wide range of applications in computer vision tasks. In the process of bounding box regression, not only can the regression state be evaluated, but also gradient propagation can be performed by calculating regression losses to accelerate convergence. Here we discuss the relationship between IoU changes and bounding box size during the regression process\cite{ref8}, analyze the inherent characteristics of bounding box regression problems and explain the rationality of the method proposed in this article.
	\par As shown in the Fig.\ref{fig_3}, Fig.3a shows the IoU Deviation curve, with the horizontal and vertical axis representing the deviation and the IoU value respectively. The three different color curves correspond to the IoU change curves of different scale bounding boxes. A, B, C, D and E correspond to 5 different positional relationships for the achors and GT boxes, where the red bounding boxes represent anchors with a length and width of 10 and the corresponding GT boxes are represented by the black bounding boxes. Fig.3b shows the ABS (Grad) Deviation curve. Unlike Fig.3a, the vertical axis in Fig.3b represents the absolute value of the IoU gradient.
	We assume that the actual bounding boxes size is 10 and the bounding boxes with sizes 8 and 12 are used as auxiliary bounding boxes. In Fig.\ref{fig_3} , A and E correspond to the regression state of low IoU samples, while B and D correspond to the regression state of high IoU samples. The following conclusions can be drawn from Fig.\ref{fig_3}.
	\par 1.Due to the scale difference between the auxiliary bounding boxes and the actual bounding boxes, the trend of IoU value change during the regression process is consistent with the trend of IoU value change of the actual bounding boxes, which can reflect the quality of the actual bounding boxes regression results.
	\par 2.For high IoU samples, the absolute value of the IoU gradient of the smaller scale auxiliary bounding boxes is greater than the absolute value of the actual bounding boxes IoU gradient.
	\par 3.For low IoU samples, the absolute value of the IoU gradient of the larger scale auxiliary bounding boxes is greater than the absolute value of the actual bounding boxes IoU gradient.
	\par Based on the above analysis, using smaller scale auxiliary bounding boxes to calculate IoU loss will help with high IoU sample regression and achieve accelerated convergence. On the contrary, using larger scale auxiliary bounding boxes to calculate IoU loss can accelerate the regression process of low IoU samples.
	\begin{figure}[!htbp]
		\subfloat[]{
			\centering
			\includegraphics[width=1.6in]{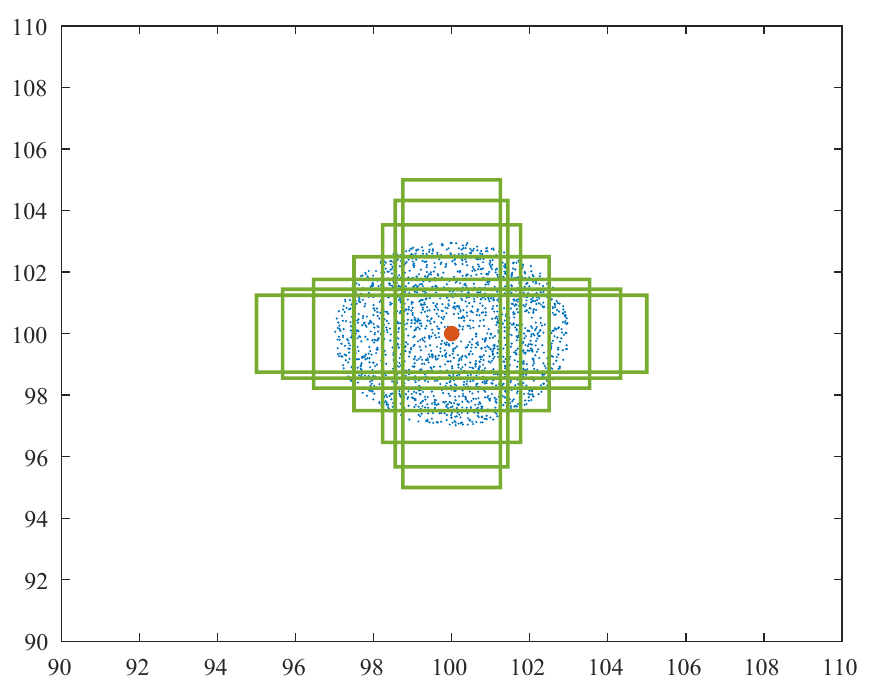}}
		\hfill
		\subfloat[]{
			\centering
			\includegraphics[width=1.6in]{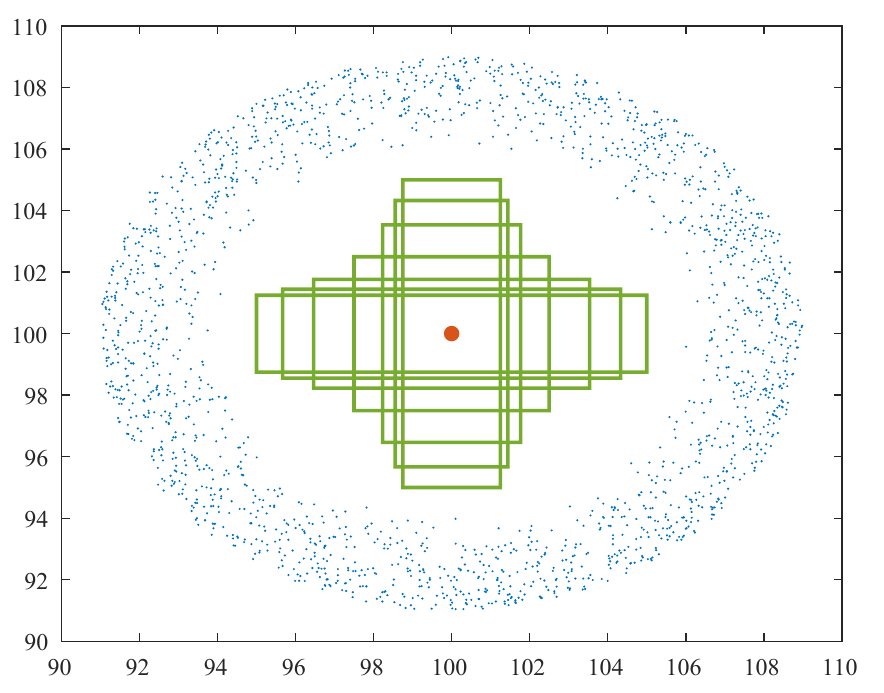}}
		\caption{Blue points denote the anchors and green bounding boxes represent the different sizes of the target box (a) high IoU regression sample (b) low IoU regression sample}
		\label{fig_6}
	\end{figure}
	\subsection{Inner-IOU Loss}
	To compensate for the weak generalization and slow convergence of existing IoU losses\cite{ref1,ref2,ref3,ref4,ref5} in different detection tasks, we propose using auxiliary bounding boxes to calculate losses and accelerate the bounding box regression process. In Inner-IoU, we introduce the scale factor ratio, which can control the scale size of the auxiliary bounding boxes. By using auxiliary bounding boxes of different scales for different datasets and detectors, the limitation of weak generalization in existing methods can be overcome. 
	\par The ground truth (GT) box and anchor are denoted as $B^{gt}$ and $B$ respectively, as illustrated in the Fig.\ref{fig_1}. The center point of the GT box and the inner GT box is represented by ($x_{c}^{gt}$,$y_{c}^{gt}$), while ($x_{c},y_{c}$) represents the center point of the anchor and the inner anchor. The width and height of the GT box are denoted as $w^{gt}$ and $h^{gt}$ respectively, while the width and height of the anchor are represented by $w$ and $h$. The variable "ratio" corresponds to the scaling factor, typically within the range of values [0.5, 1.5].
	\begin{figure*} [h]
		\centering
			\includegraphics[scale=0.21]{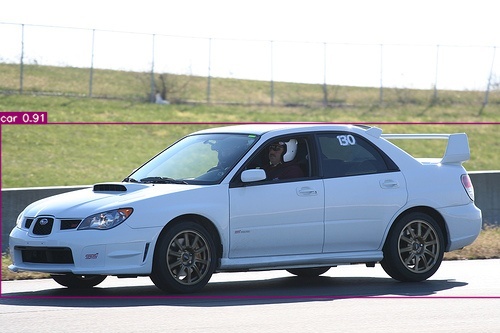}
		\hspace{0.1cm}
		\vspace{0.4cm}
			\includegraphics[scale=0.21]{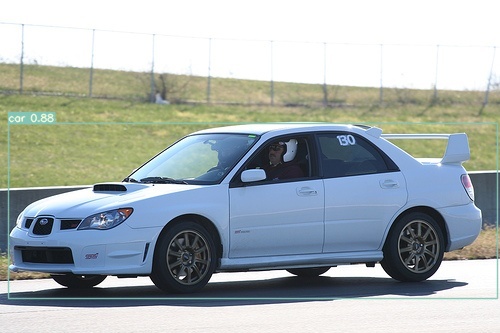}
		\hspace{0.1cm}
			\includegraphics[scale=0.21]{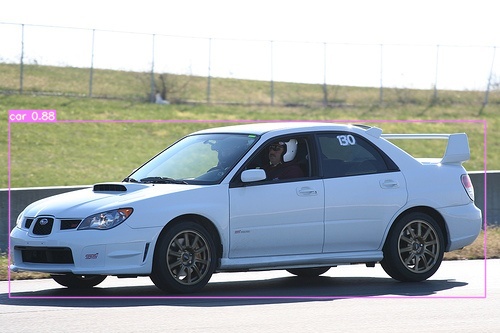}
		\hspace{0.1cm}
			\includegraphics[scale=0.21]{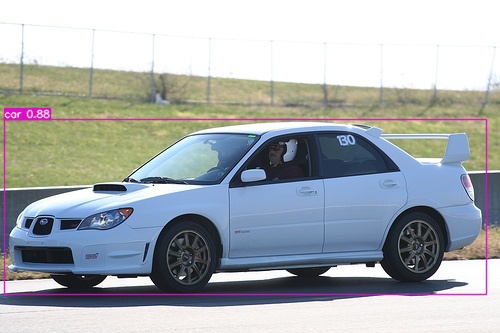}
		\\
		\centering
			\includegraphics[scale=0.21]{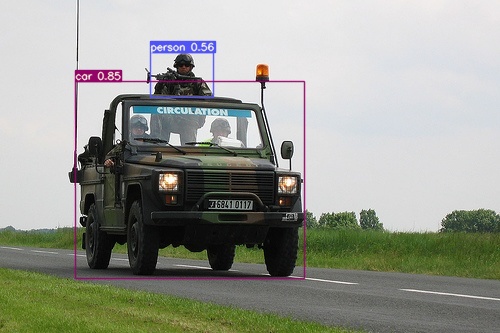}
		\hspace{0.015cm}
			\includegraphics[scale=0.21]{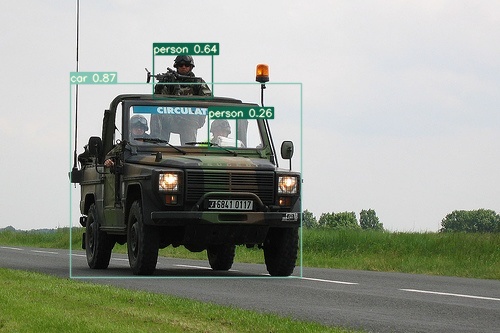}
		\hspace{0.1cm}
			\includegraphics[scale=0.21]{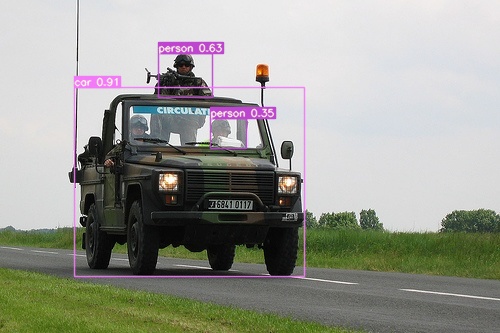}
		\hspace{0.1cm}
			\includegraphics[scale=0.21]{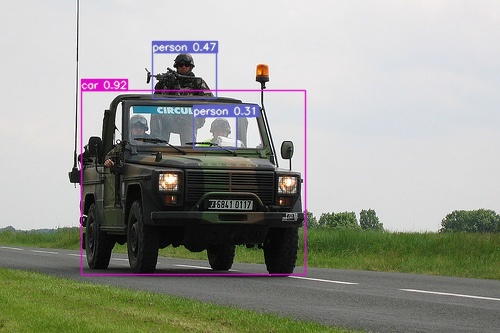}
		\caption{Detection examples on the test set of PASCAL VOC 2007 using YOLOv7-tiny by  $L_{SIoU}$ and $L_{Inner-SIoU}$. From left to right, they represent the SIoU method, Inner-SIoU (ratio=0.7), Inner-SIoU (ratio=0.75) and Inner-SIoU (ratio=0.8). }
		\label{fig_5} 
	\end{figure*}
		\begin{figure*}[!htbp]
		\subfloat[]{	
			\centering
			\includegraphics[width=3in]{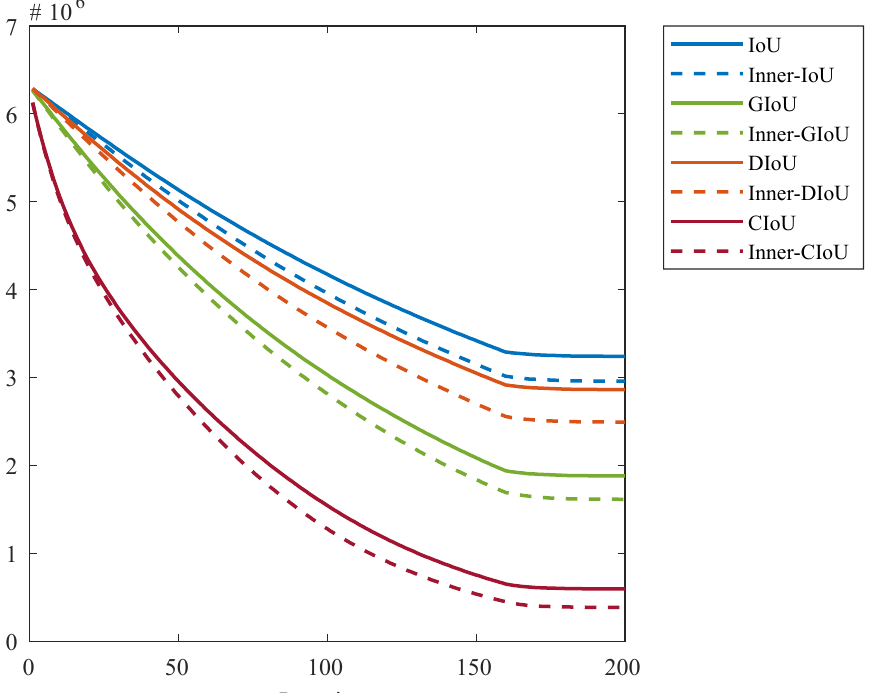}}
		\hspace{2cm}
		\subfloat[]{
			\centering
			\includegraphics[width=3in]{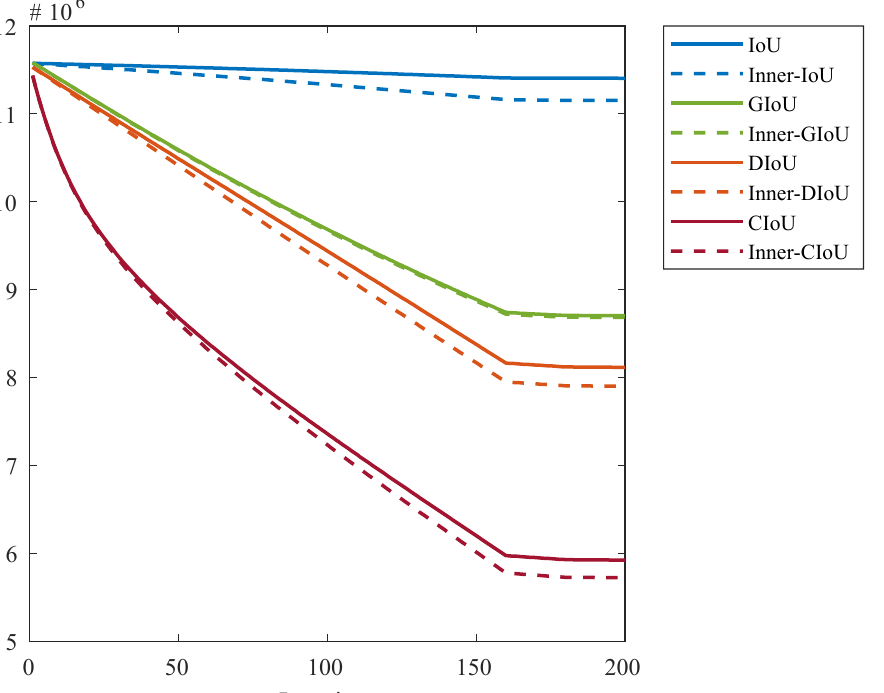}
		}
		\caption{Regression error of Inner-IoU method and several existing methods at iteration. 
		(a) high IoU sample results(ratio=0.8) (b) low IoU sample results(ratio=1.2)}
		\label{fig_4}
	\end{figure*}
	\begin{equation}
		b_{l}^{gt}=x_{c}^{gt}-\frac{w^{gt}*ratio}{2},b_{r}^{gt}=x_{c}^{gt}+\frac{w^{gt}*ratio}{2}
	\end{equation}
	\begin{equation}
		b_{t}^{gt}=y_{c}^{gt}-\frac{h^{gt}*ratio}{2},b_{b}^{gt}=y_{c}^{gt}+\frac{h^{gt}*ratio}{2}
	\end{equation}
	\begin{equation}
		b_{l}=x_{c}-\frac{w*ratio}{2},b_{r}=x_{c}+\frac{w*ratio}{2}
	\end{equation}
	\begin{equation}
		b_{t}=y_{c}-\frac{h*ratio}{2},b_{b}=y_{c}+\frac{h*ratio}{2}
	\end{equation}
	
	\begin{equation}
	\begin{aligned}
			inter=(min(b_{r}^{gt},b_{r})-max(b_{l}^{gt},b_{l}))*\\
			(min(b_{b}^{gt},b_{b})-max(b_{t}^{gt},b_{t}))
	\end{aligned}
	\end{equation}
	\begin{equation}
		union=(w^{gt}*h^{gt})*(ratio)^{2}+(w*h)*(ratio)^{2}-inter
	\end{equation}
	\begin{equation}
		IoU^{inner}=\frac{inter}{union}
	\end{equation}
	\par Inner-IoU loss inherits some of the characteristics of IoU loss, while also having its own characteristics. The range of values for Inner-IoU loss, like IoU loss, is [0,1]. Because there is only a difference in scale between the auxiliary bounding boxes and the actual bounding boxes, the calculation method of the loss function is the same and the Inner-IoU Deviation curve is similar to the IoU Deviation curve.
	\par Compared with IoU loss, when the ratio is less than 1 and the auxiliary bounding boxes size is smaller than the actual bounding boxes, the effective range of regression is smaller than IoU loss, but the absolute value of the gradient is greater than the gradient obtained from IoU loss, which can accelerate the convergence of high IoU samples. On the contrary, when the ratio is greater than 1, the larger scale auxiliary bounding boxes expand the effective range of regression and enhance effect for low IoU samples regression. Applying Inner-IoU loss to the existing IoU based bounding box regression loss function, $L_{Inner-IoU}$, $L_{Inner-GIoU}$, $L_{Inner-DIoU}$, $L_{Inner-CIoU}$, $L_{Inner-EIoU}$ and $L_{Inner-SIoU}$ are as follows:
	\begin{equation}
		L_{Inner-IoU}=1-IoU^{inner}
	\end{equation}
	\begin{equation}
		L_{Inner-GIoU}=L_{GIoU}+IoU-IoU^{inner}
	\end{equation}
	\begin{equation}
		L_{Inner-DIoU}=L_{DIoU}+IoU-IoU^{inner}
	\end{equation}
	\begin{equation}
		L_{Inner-CIoU}=L_{CIoU}+IoU-IoU^{inner}
	\end{equation}
	\begin{equation}
		L_{Inner-EIoU}=L_{EIoU}+IoU-IoU^{inner}
	\end{equation}
	\begin{equation}
		L_{Inner-SIoU}=L_{SIoU}+IoU-IoU^{inner}
	\end{equation}
	\section{Experiments}
	\subsection{Simulation Experiment}
		\begin{figure*} [t!]
		\subfloat[]{
			\centering
			\includegraphics[scale=0.4]{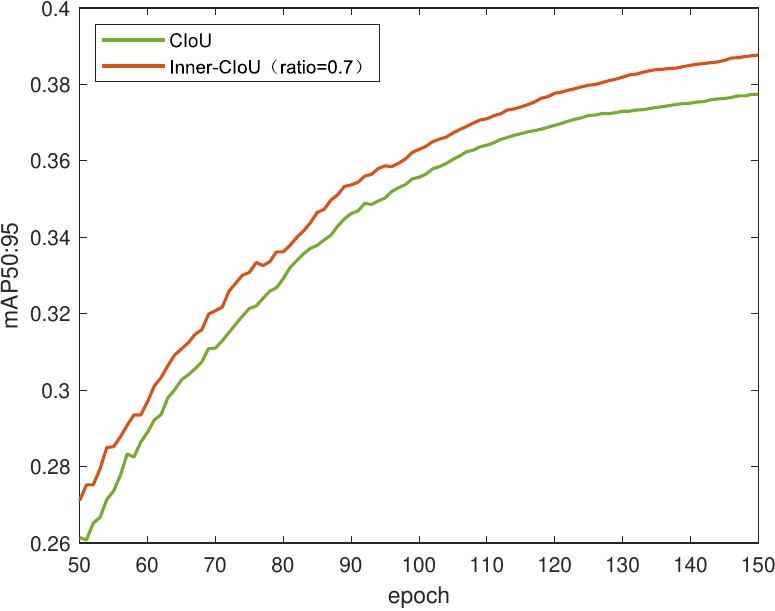}}
		\hspace{.15in}
		\subfloat[]{
			\centering
			\includegraphics[scale=0.4]{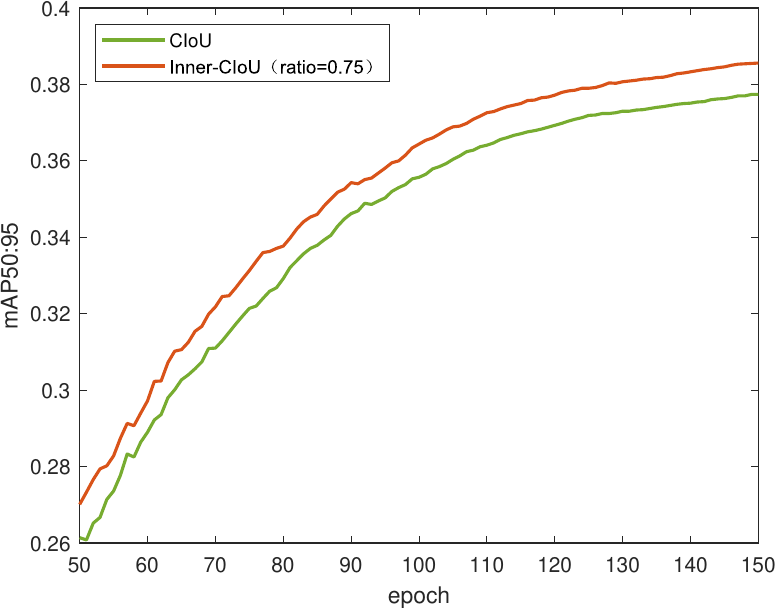}}
		\hspace{.15in}
		\subfloat[]{
			\centering
			\includegraphics[scale=0.4]{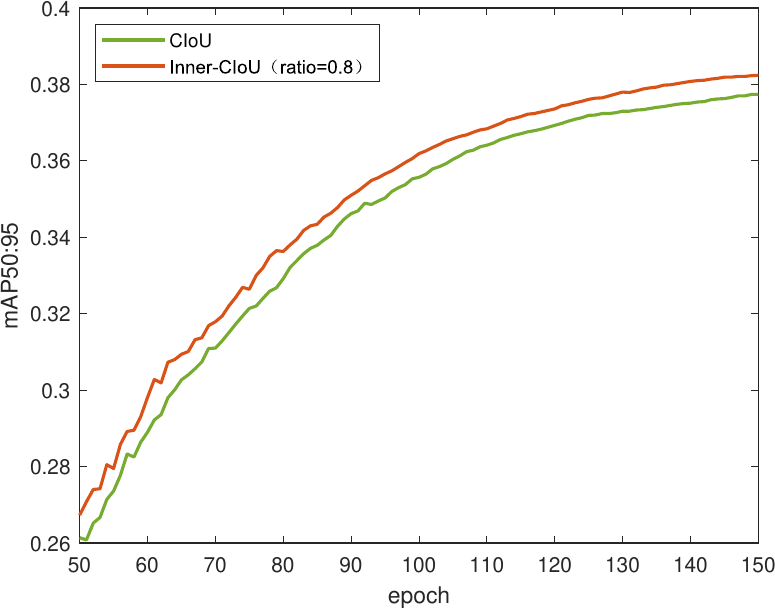}}
		\hspace{.15in}
		\\
		\subfloat[]{
			\centering
			\includegraphics[scale=0.4]{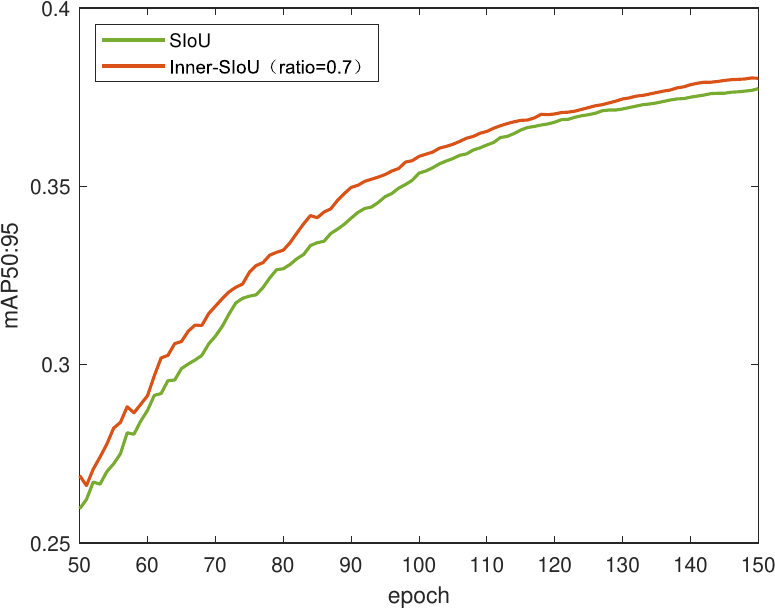}}
		\hspace{.15in}
		\subfloat[]{
			\centering
			\includegraphics[scale=0.4]{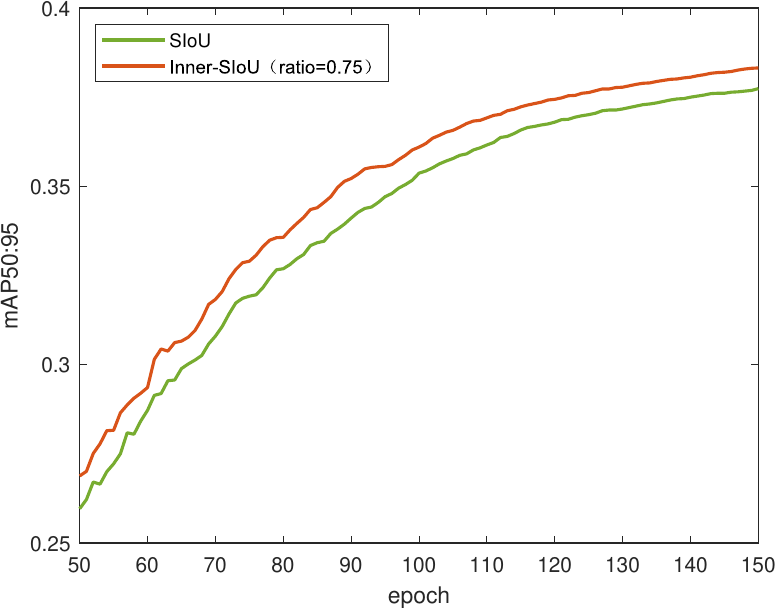}}
		\hspace{.15in}
		\subfloat[]{
			\centering
			\includegraphics[scale=0.4]{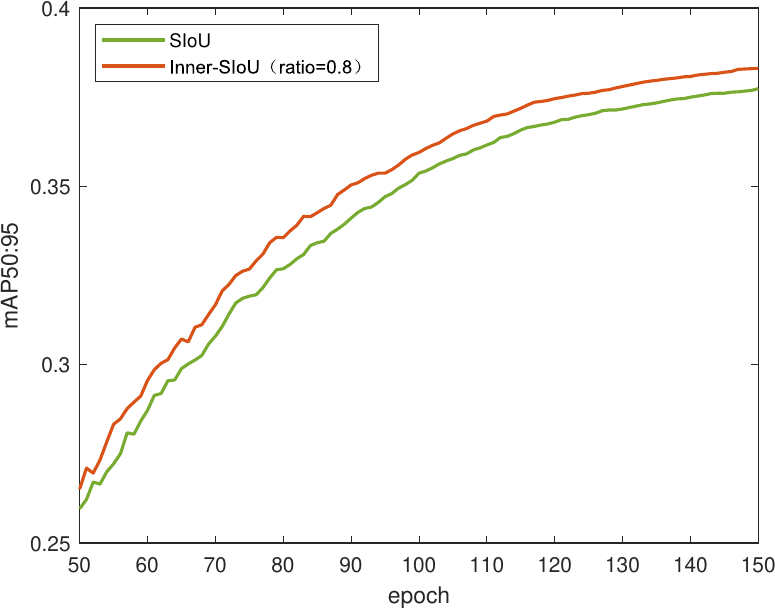}}
		\caption{Performance of CIoU and SIoU losses with different ratios.} 
		\label{fig_7} 
	\end{figure*}
	\par As shown in Fig.\ref{fig_6}, this article analyzes the bounding box regression process in two different scenarios through simulation experiments. In Fig.5a and Fig.5b, seven different green bounding boxes are set as target boxes, and the center point of the target boxes is set to (100,100), with ratios of 1:4, 1:3, 1:2, 1:1, 2:1, 3:1, and 4:1. In Fig.5a anchor boxes are randomly assigned 2000 points, with their position distribution centered on (100,100) and a radius of 3. For scale of each point, the area of anchor boxes are set as 0.5, 0.67, 0.75, 1, 1.33, 1.5 and 2. For a given point and scale, 7 aspect ratios are adapted, ie., following the same setting with target boxes (ie., 1:4, 1:3, 1:2, 1:1, 2:1,3:1, and 4:1), Fig.5b is different from Fig.5a in terms of anchor distribution, with its position distribution centered on (100,100) and a radius of 6 to 9. The dimensions and aspect ratios are the same as Fig.5a. Overall, in each experiment, 2000 × 7 × 7 anchor boxes should be fitted to each target box. To sum up, there are total 686000=7 × 7 × 7 × 2000 compression cases.
	\par The results of the simulation experiment are shown in the Fig.\ref{fig_4}, where Fig.7a represents the convergence results under the scenario of high IoU regression samples. To accelerate the regression of high IoU samples, the scale factor ratio is set to 0.8. The convergence results in the low IoU regression sample scenario are shown in Fig.7b, with the ratio set to 1.2. It can be seen that the convergence speed of our method represented by the dashed line in the figure is better than the existing methods.
	\subsection{Comparison Experiments}
	\par \textbf {YOLOv7 on PASCAL VOC}
	\par The experiment compared the CIoU\cite{ref3} method and SIoU\cite{ref5} method, using YOLOv7-tiny\cite{ref13} as the detector, VOC2007 trainval  and VOC2012 trainval as the training set and VOC2007 test as the test set\cite{ref20}. The training set consists of 16551 images, while the test set consists of 4952 images with 20 categories. We trained 150 epochs on the training set to demonstrate the superiority of our method. We visualize the training process of the proposed method and the original method, as shown in the Fig.\ref{fig_7}. Fig.8a, Fig.8b, and Fig.8c show the training process curves of CIoU and Inner-CIoU, with corresponding ratios of 0.7, 0.75, and 0.8, respectively. Fig.8d, Fig.8e and Fig.8f are the training process curves of SIoU and Inner-SIoU at ratios of 0.7, 0.75, and 0.8, respectively. In the Fig.\ref{fig_7}, the orange curve represents the method proposed in this paper, while the existing methods are represented by the green curve. It is clearly to see that the method proposed in this paper outperforms existing methods in the training process, ranging from 50 to 150 epochs.
	\par The experiment results of the comparative experiment on the test set are shown in the TABLE.\ref{tab:mytable1}. It can be seen that the detection effect has been improved after the application of the method in this article, with an increase of more than 0.5\% in AP50 and mAP50:95. Fig.\ref{fig_2} and Fig.\ref{fig_5} show a comparison of the detection samples. It can be seen from the figures that compared with existing methods, the proposed method has more accurate positioning and fewer false detections and missed detections.
	\newline
	\begin{table}[h]
		\centering
		\begin{tabular}{ccc}
			\toprule 
			& $AP_{50}$ & $mAP_{50:95}$ \\
			\midrule 
			CIoU & 63.60 & 37.64 \\
			Inner-CIoU(ratio=0.70) & 64.44(+0.84) & 38.38(+0.74) \\
			Inner-CIoU(ratio=0.75) & 64.20(+0.60) & 38.25(+0.61) \\
			Inner-CIoU(ratio=0.80) & 64.33(+0.73) & 38.30(+0.66) \\
			\midrule 
			SIoU & 63.38 & 37.31 \\
			Inner-SIoU(ratio=0.70) & 63.98(+0.60) & 38.06(+0.75) \\
			Inner-SIoU(ratio=0.75) & 64.36(+0.98) & 38.52(+1.21) \\
			Inner-SIoU(ratio=0.80) & 64.01(+0.63) & 37.98(+0.67) \\
			\bottomrule 
		\end{tabular}
		\caption{The performance of CIoU and SIoU losses(ratio between 0.7 and 0.8).}
		\label{tab:mytable1}
	\end{table}
	\newline
	\newline
	\textbf {YOLOv5 on AI-TOD}
	\par To demonstrate the generalization ability of the proposed method, we conducted comparative experiments using the YOLOv5s detector on the AI-TOD dataset\cite{ref21}, using SIoU\cite{ref5} as the comparison method.
	
	AI-TOD includes 28036 aerial images, 8 types of targets, and 700621 object instances, with 14018 images as the training set and the other 14018 images as the test set. Compared with existing target detection task datasets, the average size of AI-TOD is 12.8 pixels, which is much smaller than other datasets. The experiment results are shown in the TABLE.\ref{tab:mytable2}
	\begin{table}[h]
		\centering
		\begin{tabular}{ccc}
			\toprule 
			& $AP_{50}$ & $mAP_{50:95}$ \\
			\midrule 
			SIoU & 42.70 & 18.06 \\
			Inner-SIoU(ratio=1.10) & 43.42(+0.72) & 17.89(-0.17) \\
			Inner-SIoU(ratio=1.13) & 43.37(+0.67) & 18.23(+0.17) \\
			Inner-SIoU(ratio=1.15) & 43.77(+1.07) & 18.23(+0.17) \\
			\bottomrule 
		\end{tabular}
		\caption{The performance of SIoU loss(ratio$>$1)}
		\label{tab:mytable2}
	\end{table}
	\par In comparative experiment 1, by setting the ratio value between 0.7 and 0.8 to be less than 1, an auxiliary border smaller than the actual bounding box was generated. The experiment results demonstrate that it can generate gain on high IoU samples. In experiment 2, when the ratio value is greater than 1, the effect of accelerating convergence for low IoU samples is achieved by generating larger auxiliary bounding boxes. In addition, Fig.\ref{fig_8} shows the comparison of detection results on the test set, and the superiority of this method can be seen through comparison.
	\section{Conclusion}
	\par In this article, we analyzed the bounding box regression process and pointed out the limitation of IoU loss, which does not have strong generalization for different detection tasks. Based on the inherent characteristics of the bounding box regression problem, we propose Inner-IoU loss, a bounding box regression loss based on auxiliary bounding boxes. It controls the generation of auxiliary bounding boxes through the scale factor ratio to calculate losses and accelerate convergence. It can be integrated into existing IoU based loss functions.
	Through a series of simulation and ablation experiments, it is verified that the proposed method is superior to existing methods. It should be noted that the method proposed in this paper is not only applicable to general detection tasks, but also performs well for detection tasks with very small targets, and the generalization of the method has been confirmed.

\newpage

\vfill

\end{document}